# The Linguistic Relevance of Quasi-Trees

Tony Kroch, University of Pennsylvania
Owen Rambow, Université Paris 7
kroch@linc.cis.upenn.edu, rambow@linguist.jussieu.fr

**Abstract**

We discuss two constructions (long scrambling and ECM verbs) which challenge most syntactic theories (including traditional TAG approaches) since they seem to require exceptional mechanisms and postulates. We argue that these constructions should in fact be analyzed in a similar manner, namely as involving a verb which selects for a "defective" complement. These complements are defective in that they lack certain Case-assigning abilities (represented as functional heads). The constructions differ in how many such abilities are lacking. Following the previous analysis of scrambling of Rambow (1994), we propose a TAG analysis based on quasi-trees.

## 1. Introduction

Vijay-Shanker (1992) discusses general requirements on the formalism used to represent syntactic competence and suggests that minimal assumptions lead to the adoption of an underspecified tree structure, which he calls a *quasi-tree*. A quasi-tree is like a tree, except that each node is broken into two parts, a bottom part and a top part. In a derivation,[1] quasi-trees are composed by performing substitutions at the quasi-nodes. Trees can interleave, but the top of a quasi-node must dominate its bottom in the final derived tree. Observe that quasi-trees represent a generalization of traditional trees, such as those of tree adjoining grammar (TAG). Adjunction in a TAG is modeled by two quasi-tree substitutions, but we can also combine two quasi-trees by performing three or more substitutions, which leads to the "interleaving" of trees. There is no longer a corresponding TAG operation.

In this paper, we reinforce Vijay-Shanker's argument in favor of quasi-trees by using specific linguist examples. Two apparently unrelated constructions, long scrambling and Exceptional Case-Marking (ECM), can be easily expressed in quasi-trees in a similar manner. We will argue that both constructions are in fact instances of a similar phenomenon, namely a verb selecting for a "defective" complement which lacks certain Case-assigning abilities.[2] Arguments of the embedded verb must therefore look to the embedding verb for their Case, giving rise to constructions in which the "domains" of two verbs, i.e. the elementary structures associated with two verbs, are interleaved. Such interleaving is naturally expressed using quasi-tree.

This paper is structured as follows. In Section 2, we present the data for the two constructions and informally introduce our syntactic analyses and highlight their commonalities. In Section 3, we present an analysis of scrambling based on quasi-trees, and then show how this analysis transfers to ECM verbs.

## 2. Data and Informal Analyses

### 2.1 Scrambling

In German, scrambling is the permutation of arguments in the so-called *Mittelfeld*, the area between the finite verb in second position and the sentence-final non-finite verbs. In "long scrambling," an argument of a verb is located to the left of an argument of a less deeply embedded verb. Long scrambling is illustrated in (1):

| | | | | | |
|---|---|---|---|---|---|
| (1) | . . . daß [den Kühlschrank]$_i$ | niemand | [t$_i$ zu reparieren] | versprochen | hat |
| | . . . that the refrigerator $_{\text{ACC}}$ | no-one $_{\text{NOM}}$ | to repair | promised | has |

[1] Quasi-trees are in fact only a data structure, and various rewriting systems can be defined based on them. We use the simplest possible system, called quasi-tree substitution grammar in (Rambow, 1994b).

[2] In (Kroch and Rambow, 1994) we extend the argument to Clitic Climbing.

Long scrambling is only licensed by certain verbs. For example, *verlangen* 'to demand' does not license long scrambling:

(2) *. . . daß den Kühlschrank niemand zu reparieren verlangte

Thus it is clear that properties of the embedding verb determine whether long scrambling of arguments of the embedded verb is possible. In the past, it has been proposed that certain verbs allow for some form of restructuring, in which the embedded clause boundary is deleted ("clause pruning") (Haegeman and van Riemsdijk, 1986), or the verbs incorporate and form a morphological unit with a merged theta-grid ("clause union") (Grewendorf, 1988), or both (Evers, 1975). We reject the notion of a special process of incorporation as unfounded and unnecessarily complex (see (Kroch and Santorini, 1991)), but adopt and adapt the notion of clause pruning, which we reinterpret in a nontransformational framework to mean that the complement clause may be (but need not be) deficient with respect to certain Case-assigning abilities. In particular, verbs that allow for long scrambling optionally select for embedded verbs in whose extended projection some (or all) Case-assigning heads are altogether absent. As a result, the arguments in question must receive Case in the matrix clause. Note that under this analysis, we can preserve the theta-assigning role of the embedded verb, thus avoiding the problems that the clause-union analysis encounters with the Projection Principle (or, put differently, we can avoid postulating an infinite lexicon).

How do these NPs get Case in the matrix clause? If we pursue this line of inquiry, we are led to dissociate Case-assigning ability and $\theta$-assignment. Thus, an argument need not receive Case from its governing verb. This approach is suggested independently by Korean data which shows that adjuncts can be assigned Case as well (Lee, 1993). (See also (Haeberli, 1993) for a similar suggestion for German.)

### 2.2 ECM Verbs

ECM verbs such as *believe* assigned accusative Case to the subject of the subordinate clause:[3]

(3) We believed him to like kidneys

In transformational analyses, such verbs have been analyzed as deleting the CP (or S′) barrier, and thus allowing accusative to be assigned under Government. In TAG-based approaches, the same intuition can be implemented by using a feature which ECM verbs assign to their complements and which these pass on to their subject, forcing accusative Case. In both approaches, these verbs require additional mechanisms. We would like to be able to reduce the difference between ECM verbs and object-control verbs to the simple fact that the former, unlike the latter, select for a clause that does not functionally license a subject (be it a nominative one or a PRO subject[4]), even though the complement verb requires one. Thus, in the case of ECM verbs, the embedded subject raises to the matrix clause to receive Case there. Under this approach, ECM verbs are very similar to raising verbs. (The term "raising-to-object" has been used in the past for ECM.) In both cases, the embedded subject is not assigned Case within its clause and "raises" for this purpose to the matrix clause.

## 3. A Quasi-Tree Analysis

### 3.1 Scrambling

Local scrambling does not pose any particular problem from the point of view of TAG, since we can simply generate all possible orderings in simple TAG trees. However, long scrambling requires additional mechanisms, as argued in (Becker et al., 1991), such as multi-component TAG. Simple TAG is *formally* inadequate. In (Rambow, 1994a), it is proposed to use tree sets in a multi-component TAG variety called V-TAG as the basic representation for the German clause, and derive both local and long-distance scrambling by the same means. (Rambow, 1994b) shows that V-TAG is in fact a notational variant of quasi-trees. The proposed analysis, in terms of quasi-trees, is shown in Figure 1. (Details are omitted.)

[3]The fact that ECM verbs allow expletive embedded subjects (*we expect there to be a riot*) shows that that element cannnot be an argument of the matrix clause (or, put differently, ECM verbs are not object-control verbs).

[4]Watanabe (1993) assimilates these two cases by proposing that PRO in fact receives a special case (PRO-case) from AgrS.

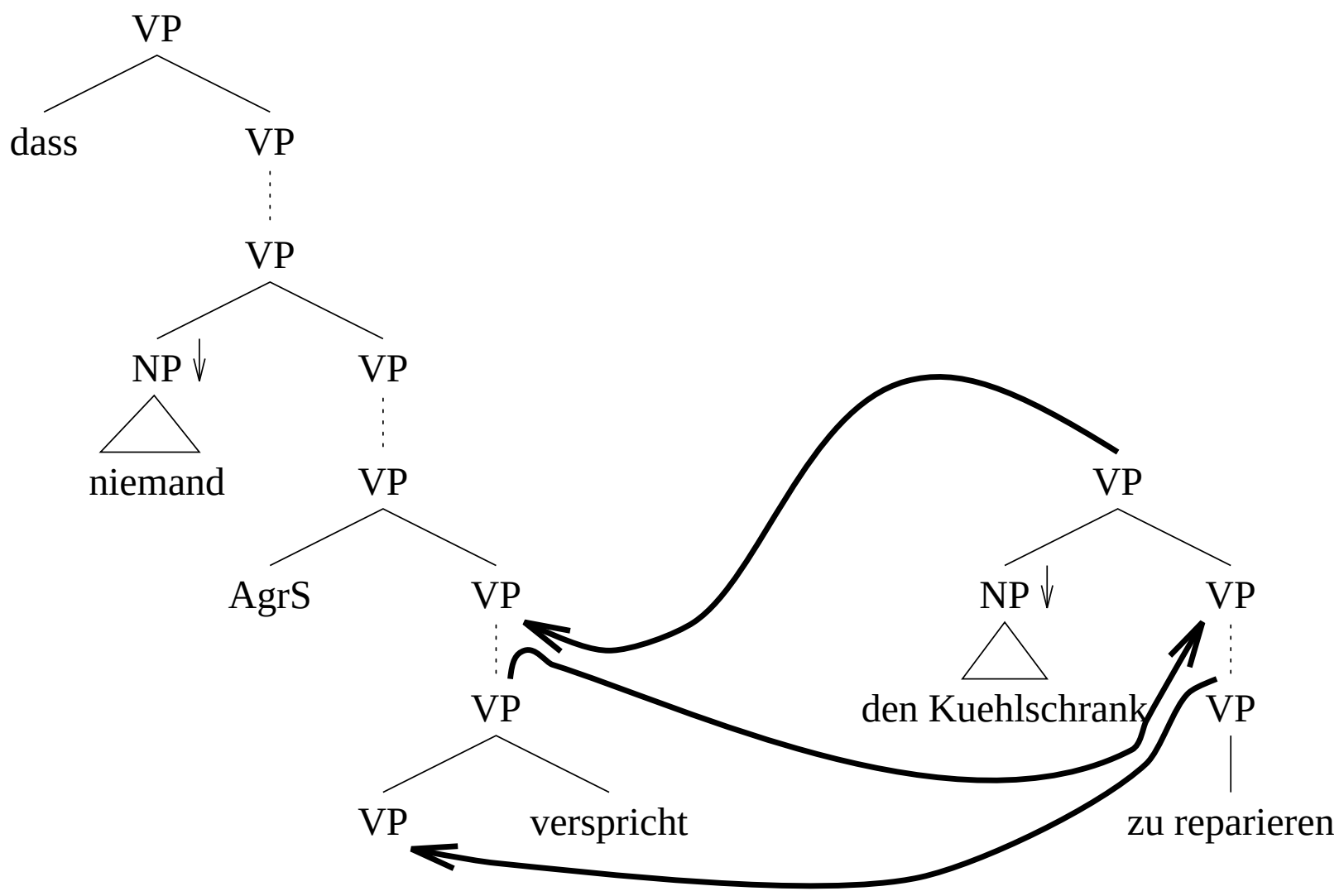

Figure 1: Quasi-tree analysis of Scrambling

### 3.2 ECM Verbs

As we have seen in Section 2.2, we want to analyze ECM verbs like raising-to-subject verbs. In the TAG analysis of raising(-to-subject) verbs (Kroch and Joshi, 1985; Frank, 1992), we perform an adjunction of the tree associated with the matrix verb after which the subject of the embedded clause is in the proper configuration in the matrix clause to receive Case. (For conceptual simplicity, we will assume that Case is uniformaly assigned under Spec-Head agreement, by an AgrS head (nominative Case) or an AgrO head (accusative case) (see (Lasnik, 1993)).) In an ECM construction, however, the embedded subject is not in sentence-initial position. Put schematically, we would like to obtain the string in (4) by performing an adjunction such that the substrings in **boldface** (the embedded clause) originate in one tree and the strings in regular typeface (the matrix clause) originate in another tree. Furthermore, we would like to have a right-branching structure. Clearly, this is impossible in a TAG.

(4) We believed **him** AgrO **to like kidneys**

The problem disappears when we use quasi-trees. We can now have the subject be part of the embedded clause, and the AgrO head be part of the matrix clause. We perform three substitutions and receive the desired right-branching structure. Observe that the only difference between ECM and control verbs is that the former, but not the latter, specify that the complement lacks an AgrS.

### 4. Conclusion

In this paper, we have suggested that two apparently different constructions should be analyzed in a similar manner: a verb selects for a complement clause which lacks certain functional projections that assign Case. The need for all embedded arguments to receive Case forces embedded arguments to interleave with Case-assigning heads from the embedding clause. If our argument is correct that this interplay of selection for a “defective” complement and Case theory is responsible for a variety of constructions, then this represents a strong argument for the use of quasi-trees, which can elegantly implement such interleaving. Furthermore, the use of quasi-trees has the advantage that the elementary structures used by these constructions differ minimally from the standard clausal projections, namely only in the presence or absence of certain Case-assigning functional projections.

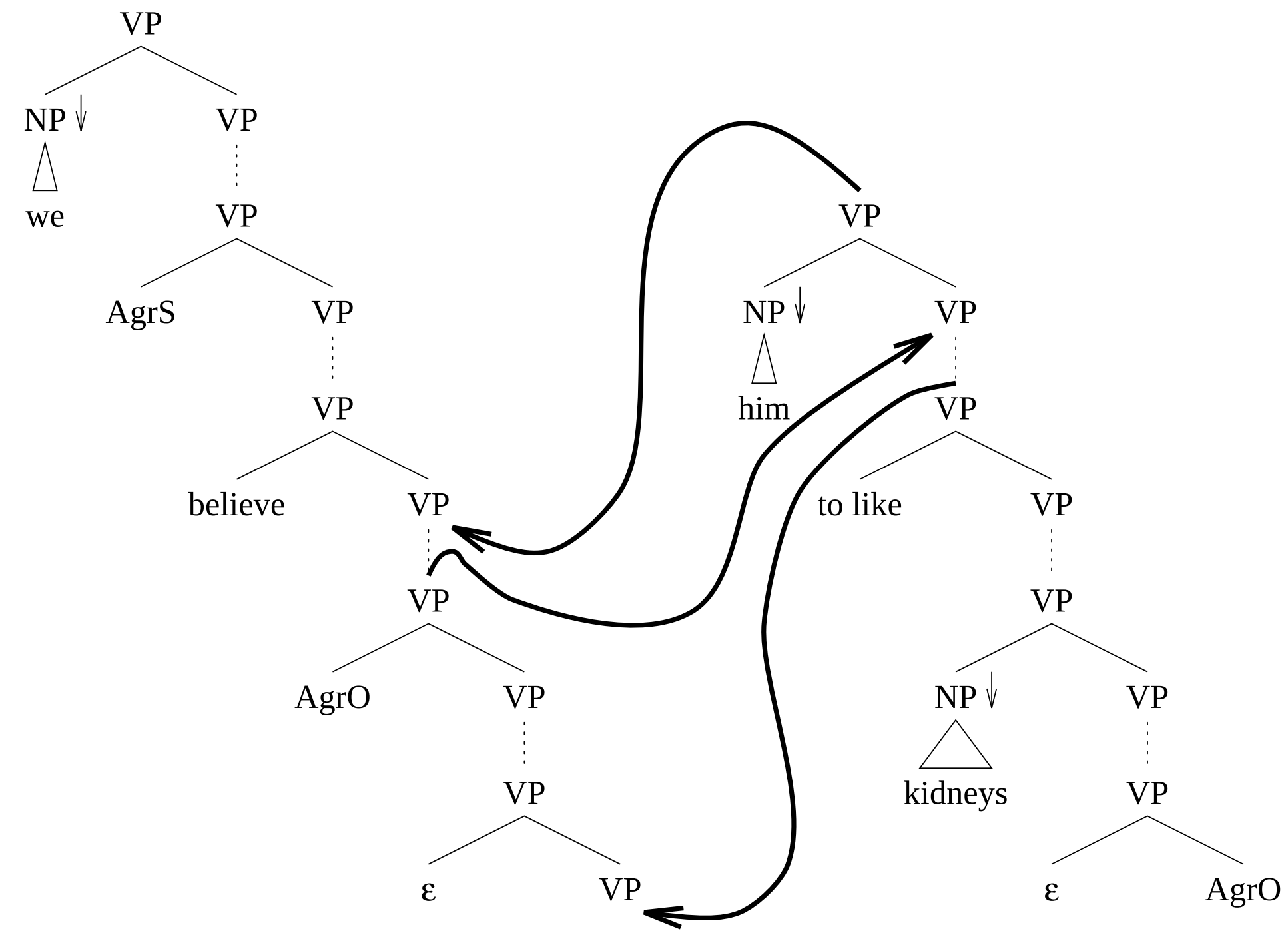

Figure 2: Quasi-tree analysis of ECM verbs